\documentclass[conference]{IEEEtran}
\IEEEoverridecommandlockouts

\newcommand\copyrighttext{%
  \footnotesize \textcopyright 2020 IEEE. Personal use of this material is permitted.  Permission from IEEE must be obtained for all other uses, in any current or future media, including reprinting/republishing this material for advertising or promotional
  purposes, creating new collective works, for resale or redistribution to servers or lists, or reuse of any copyrighted component of this work in other works.}
\newcommand\arxivcopyrightnotice{%
\begin{tikzpicture}[remember picture,overlay]
\node[anchor=south,yshift=10pt] at (current page.south) {\fbox{\parbox{\dimexpr\textwidth-\fboxsep-\fboxrule\relax}{\copyrighttext}}};
\end{tikzpicture}%
}

\usepackage{textcomp}
\usepackage{hyperref}
\usepackage{tikz}
\usepackage{cite}
\usepackage{amsmath,amssymb,amsfonts}
\usepackage{algorithmic}
\usepackage{graphicx}
\usepackage{textcomp}
\usepackage{xcolor}
\usepackage{subcaption}
\def\BibTeX{{\rm B\kern-.05em{\sc i\kern-.025em b}\kern-.08em
    T\kern-.1667em\lower.7ex\hbox{E}\kern-.125emX}}
\begin{document}

\bibliographystyle{IEEEtran}

\title{Inter- and Intra-domain Knowledge Transfer for Related Tasks in Deep Character Recognition
\thanks{This work is based on the research supported in part by the National Research Foundation of South Africa (Grant Number: 17808).}
}

\author{\IEEEauthorblockN{Nishai Kooverjee}
\IEEEauthorblockA{\textit{School of Computer Science} \\
\textit{and Applied Mathematics} \\
\textit{University of the Witwatersrand}\\
Johannesburg, South Africa \\
nishai.kooverjee1@students.wits.ac.za}
\and
\IEEEauthorblockN{Steven James}
\IEEEauthorblockA{\textit{School of Computer Science} \\
\textit{and Applied Mathematics} \\
\textit{University of the Witwatersrand}\\
Johannesburg, South Africa \\
steven.james@wits.ac.za}
\and
\IEEEauthorblockN{Terence van Zyl}
\IEEEauthorblockA{\textit{School of Computer Science} \\
\textit{and Applied Mathematics} \\
\textit{University of the Witwatersrand}\\
Johannesburg, South Africa \\
terence.vanzyl@wits.ac.za}
}


\maketitle

\arxivcopyrightnotice

\IEEEpubidadjcol

\begin{abstract}
Pre-training a deep neural network on the ImageNet dataset is a common practice for training deep learning models, and generally yields improved performance and faster training times. The technique of pre-training on one task and then retraining on a new one is called transfer learning. In this paper we analyse the effectiveness of using deep transfer learning for character recognition tasks. We perform three sets of experiments with varying levels of similarity between source and target tasks to investigate the behaviour of different types of knowledge transfer. We transfer both parameters and features and analyse their behaviour. Our results demonstrate that no significant advantage is gained by using a transfer learning approach over a traditional machine learning approach for our character recognition tasks. This suggests that using transfer learning does not necessarily presuppose a better performing model in all cases.
\end{abstract}

\begin{IEEEkeywords}
deep learning, transfer learning, knowledge transfer, character recognition
\end{IEEEkeywords}

\section{Introduction} \label{introduction}
Learning to drive a car makes learning to drive a truck easier, and knowing how to speak Spanish makes learning Portuguese easier. Humans find it easier to learn something new when it is similar to something they already know. With this intuition, can a machine learner exploit previous experience or information for a new task?

This is the motivation behind transfer learning. We expect that, like humans, learning algorithms may  benefit from knowledge transferred from similar or related tasks. This poses many questions such as: How can this transfer be implemented? What makes tasks similar? Does knowledge from dissimilar tasks transfer well or even at all? How much knowledge should be transferred?

Traditional machine learning models are formulated with the assumption that the training and testing data are independent and identically distributed (i.i.d), whereas transfer learning does not make this assumption. Thus the distributions of the training and testing data may be different which allow for different domains or tasks to be associated with each data set.

Owing to the recent successes in deep learning \cite{alexnet, googlenet, ciresandeepbig}, this research focuses on transfer learning using deep artificial neural networks. Large, very deep models have recently achieved valuable results on a variety of tasks, but require a massive amount of training data, time, and computing resources. Transfer learning allows us to leverage existing deep models to train new deep models with a more moderate amounts of computing resources and training data in a reasonable amount of time \cite{betterImagenet}. This is achieved by transferring knowledge learnt for one task (\textit{source} task) to a different task (\textit{target} task).

Deep convolutional neural networks have been particularly successful for many complex tasks: mainly within the field of computer vision \cite{conv}. A common approach used for image classification tasks is to pre-train convolutional neural networks (CNNs) on the ImageNet dataset, before training it for the actual task. As a result, ImageNet's transferability has been well researched \cite{betterImagenet,imagenet_good}. 

This work focuses on knowledge transfer within the domain of character recognition. A commonly used dataset in deep learning is MNIST, an image collection of handwritten digits, and is often used as a baseline for comparing network architectures \cite{mnist}. We perform transfer learning experiments using MNIST, as well as the NIST Special Database 19, which is a larger dataset containing images of handwritten digits as well as characters \cite{nist}.

Our work aims to investigate whether transfer helps improve a network's convergence as well as its performance on the target task. Does transfer between related tasks work better than transfer between dissimilar tasks? Should we be transferring network parameters (\textit{fine-tuned} transfer) or learned features (\textit{frozen-weight} transfer)?

We perform three sets of transfer experiments, and compute performance metrics across all three sets over the course of training. Each experiment varies in how related the source and target tasks are, in order to quantify how task similarity affects transfer. Within each set \textit{control}, \textit{frozen-weight} and \textit{fine-tuned} networks are trained, analysed and compared. This allows us to compare how both \textit{parameter transfer} and \textit{feature transfer} perform for the same task.

We observe that \textit{fine-tuned} network performance closely resembles that of randomly-initialised networks. Performance for \textit{frozen-weight} transfer, where the network acts as a fixed feature extractor, is affected by the relatedness of the source and target tasks. We demonstrate that a transfer learning approach does not necessarily provide better starting points or improved accuracy for a network on a specific task.


\section{Background and related work} \label{background}
\subsection{Transfer Learning}
We are interested in transferring knowledge from one problem to improve performance for a new problem. To formalise this we introduce notation and definitions first given by Pan and Yang \cite{panyang}:

First we define a domain:
a domain $\mathcal{D}$ is characterised by a feature space $\mathcal{X}$, and a marginal probability distribution $P(X)$ where $X=\{\mathbf{x}_1,\dots, \mathbf{x}_n\} \in \mathcal{X}$. $\mathcal{X}$ is the space of all possible feature vectors, and $X$ is a particular learning sample. 

Given a specific domain $\mathcal{D}=\{\mathcal{X}, P(X)\}$, a task $\mathcal{T}$ is defined as consisting of a label space $\mathcal{Y}$, and a predictive function $f(\cdot)$ which is used to predict the corresponding label for a new instance $\mathbf{x}$ (i.e. $f(\mathbf{x}) := P(y|\mathbf{x})$). This is denoted as $\mathcal{T}=\{\mathcal{Y}, f(\cdot)\}$.

For simplicity, it is assumed that there is only one source domain, $\mathcal{D}_S$, and one target domain, $\mathcal{D}_T$. The \textit{source domain data} is denoted as $\mathcal{D}_S = \{(\mathbf{x}_{S_1}, y_{S_1}),\dots,(\mathbf{x}_{S_n}, y_{S_n})\}$, where $\mathbf{x}_{S_i} \in \mathcal{X}_S$ and $y_{S_i} \in \mathcal{Y}_S$ are the data instance and its corresponding label respectively. Similarly, we denote the \textit{target domain data} as $\mathcal{D}_T = \{(\mathbf{x}_{T_1}, y_{T_1}),\dots,(\mathbf{x}_{T_n}, y_{T_n})\}$, where $\mathbf{x}_{T_i} \in \mathcal{X}_T$ and $y_{T_i} \in \mathcal{Y}_T$.

Now, transfer learning is formally defined:

\textit{Given a source domain $\mathcal{D}_S$ and target domain $\mathcal{D}_T$, with corresponding source and target tasks $\mathcal{T}_S$ and $\mathcal{T}_T$, transfer learning aims to improve the predictive function $f(\cdot)$ in $\mathcal{D}_T$, by using knowledge from $\mathcal{D}_S$ and $\mathcal{T}_S$, where $\mathcal{D}_S \neq\mathcal{D}_T$ or $\mathcal{T}_S \neq\mathcal{T}_T$.}

This definition states that transfer learning aims to generalise from knowledge learnt in one domain for a specific task, to another domain or another task. When the source domain and target domain are the same, i.e. $\mathcal{D}_S=\mathcal{D}_T$, and the learning tasks are the same, i.e. $\mathcal{T}_S=\mathcal{T}_T$, this amounts to a traditional machine learning problem.

The tasks learnt from source domains $\{\mathcal{D}_{S_1},\dots,\mathcal{D}_{S_n}\}$, namely $\{\mathcal{T}_{S_1},\dots,\mathcal{T}_{S_n}\}$, are called support tasks by Thrun and O'Sullivan \cite{thrunclustering}, who note that the existence of these support tasks differentiates a transfer learning problem from a traditional machine learning problem.

\subsection{Convolutional Neural Networks}
Artificial neural networks (ANNs) work by combining artificial \textit{neurons} with other artificial \textit{neurons} in a non-linear manner, so that complex representations can be learnt to achieve some task. Conventionally, these neurons are arranged in \textit{layers}. The first layer represents the network’s \textit{input} and the last layer represents the network’s \textit{output}. Any layers between the input and output layers are called \textit{hidden layers}. The presence of hidden layers is what is referred to as a \textit{deep neural network}.

Convolutional neural networks (CNNs), are a specific class of ANNs created specifically for image processing tasks. CNNs learn features to recognise shapes with a high degree of invariance to translation, scaling, skewing, and other types of distortion \cite{haykin-nn}.

CNNs achieve this by including specific types of layers with different respective purposes:
\begin{itemize}
	\item \textit{Feature extraction:} Each neuron takes inputs only from a local receptive field in the previous layer, thus extracting its local features.
	\item \textit{Feature mapping:} Each layer of the network consists of multiple feature maps, where individual neurons are constrained to share the same weights. This reduced the number of parameters, and allows a convolution with a small kernel to be performed on each feature map.
	\item \textit{Subsampling:} Each convolutional layer is followed by a layer that performs local averaging and subsampling. This reduces the feature map's sensitivity to shifts and distortion. This allows a reduction in the feature map's size, so more specific features can be learnt.
\end{itemize}

\begin{figure}[tb]
    \centering
    \includegraphics[width=\columnwidth]{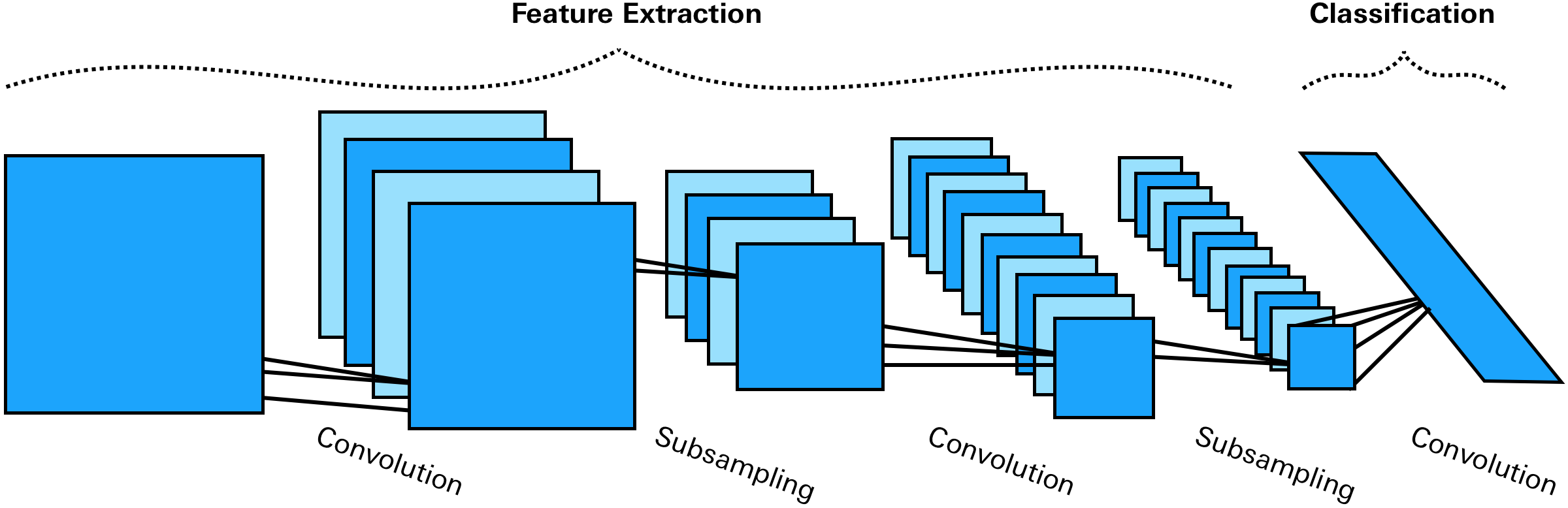}
    \caption{Example CNN architecture with alternating \textit{convolution} and \textit{subsampling} layers \cite{haykin-nn}.}
    \label{fig:cnn}
\end{figure}

The final layer of a CNN used for classification is a \textit{fully-connected} layer where the final feature maps are connected to the output neurons representing each class. Thus a CNN can be thought of as a feature extractor (the stacked \textit{convolution} and \textit{subsampling} layers that produce \textit{feature maps}) and a classifier (the \textit{fully-connected} final layer). This is shown in Fig. \ref{fig:cnn}. For transfer learning, we can transfer the feature extractor, and retrain the classifier on the new task.

\subsection{Related Work}
\subsubsection{Deep Transfer Learning}
Our work focuses on deep transfer learning since deep learning algorithms discover good multi-level representations in input distributions: with higher-level learned features defined in terms of lower-level features. Deep learning is well suited for transfer learning because it focuses on learning these representations, and in particular ``abstract'' representations that disentangle the factors of variation in the input \cite{bengiodeep}.

Clune, Bengio, \textit{et. al } \cite{howtransferable} investigate how transferable the features learnt by a deep neural network are. It is noted that CNNs learn the same features in the first or second layer, (which can be understood as Gabor filters and colour blobs). These  first-layer features are called \textit{general}, and the last layer features of a trained network are called \textit{specific}, as they are task-dependent. Thus, there must be a transition between \textit{general} and \textit{specific} features somewhere in the network.

The paper investigates transferring different numbers of layers from the pre-trained model. The transferred weights were either \textit{frozen}, where the transferred weights could not be updated, or \textit{fine-tuned}, where the transferred weights are re-learnt for the new task. It was found that after transferring more than two layers and keeping the weights \textit{frozen}, transfer accuracy decreased, indicating that \textit{task-specific} features were learnt beyond the first two layers. \textit{Fine-tuning} the weights not only avoided this drop in performance, but also improved generalisation when compared with networks trained directly on the target task.

\subsubsection{ImageNet}
A common approach for image classification tasks is to pre-train CNNs on ImageNet, a massive image dataset with hand-annotated labels and a hierarchical structure \cite{imagenet}. Following this, the network is trained on the actual task, and improved training performance is usually achieved \cite{betterImagenet, imagenet_good, howtransferable}. This begs the question: why does pre-training on ImageNet achieve this improved performance?

Kornblith, \textit{et al.} \cite{betterImagenet} investigate the hypothesis that better ImageNet performance implies better transfer performance for vision tasks. The research analyses both classification architectures and features of ImageNet models and their effects on transferability. Their results conclude that  ImageNet accuracy is very highly correlated with the \textit{fine-tuned} transfer accuracy.  However, ImageNet accuracy did not necessarily predict accuracy when fine-tuning weights, although models converge to their best accuracy levels much quicker. On average there was a 17-fold speedup over random network initialisation.

Why does the correlation between ImageNet performance and transfer performance exist? Huh, \textit{et al}. \cite{imagenet_good} investigate which aspects of ImageNet are most critical for learning good general-purpose features. Insights gained include: fine-grained recognition does not appear to be essential for learning good features to transfer\footnote{This is recognising small semantic differences between subclasses in the ImageNet hierarchy. \textit{e.g.} pugs vs bulldogs. This is as opposed to coarse-grained recognition \textit{e.g.} cats vs dogs.}, CNNs that are trained for coarse-grained classification appear to induce features that are useful in discrimination between fine-grained categories, and subclasses with common visual or spatial structures help the CNN learn more general features and thus improves transferability.

\subsubsection{Deep Transfer Learning for Character Recognition}
Cireşan, \textit{et al}. \cite{ciresanchinese} analyse transfer learning for various character recognition tasks using deep neural networks. Datasets of handwritten Latin and Chinese characters were used. Three sets of transfer experiments were performed: \textit{Digits to uppercase characters}, \textit{Chinese characters to uppercase letters} and \textit{Uppercase letters to Chinese characters}. The number of layers transferred were varied, and in all cases, better performance and improved learning convergence was achieved when compared with randomly initialised networks.

\section{Experimental Methodology} \label{methodology}
Our work aims to investigate similar sets of deep transfer learning problems within digit and character classification. The datasets we work with are \textit{MNIST} and the \textit{NIST Standard Database 19}  (henceforth referred to as \textit{NIST}) \cite{nist}. The \textit{MNIST} dataset contains grayscale images of handwritten digits with size 28x28 pixels. \textit{NIST} contains images of handwritten digits as well as both uppercase and lowercase Latin characters. The \textit{NIST} images are of size 128x128 pixels. Due to the difference in size, the MNIST images are upsampled to 128x128 when used.

Specifically, we use the ``by\_class'' partition of \textit{NIST}, where images are grouped by their label: the ASCII value for the corresponding character. We use the ``train'' directory as the training set, and the ``hsf\_4'' directory as the testing set, as is advised \cite{nist}.

\subsection{Experiments}
We perform three sets of transfer learning experiments on digit and character classifiers using CNNs: specifically, the chosen network architecture is \textit{ResNet-18} \cite{resnet}. The \textit{source} and \textit{target} tasks for each set are given in Table \ref{tab:expsets}.

\begin{table}[h]
    \centering
    \caption{Sets of experiments with source and target tasks.}
    \begin{tabular}{|c|c c|}
    \hline
     Exp. set & Source task $(\mathcal{T}_S)$ & Target task $(\mathcal{T}_T)$ \\ 
     \hline
     1 & \textit{MNIST} & \textit{NIST} Digits \\  
     2 & \textit{MNIST} & \textit{NIST} Digits + Characters \\ 
     3 & \textit{NIST} Digits & \textit{NIST} Characters \\
     \hline
    \end{tabular}
    \label{tab:expsets}
\end{table}

These three experiment sets were selected as each one corresponds to a different problem involving knowledge transfer. For \textit{MNIST to NIST Digits}, the source and target domains (hand-written character recognition) and tasks (10-digit classification) are the same, i.e. $\mathcal{D}_S = \mathcal{D}_T$ and $\mathcal{T}_S = \mathcal{T}_T$. This amounts to a normal machine learning problem, where knowledge transfer is investigated between different datasets, i.e. different samples from the same feature space $\mathcal{X}_{\text{digits}}$. For both \textit{MNIST to NIST Digits + Characters} and \textit{NIST Digits to NIST Characters}, we have $\mathcal{D}_S \neq \mathcal{D}_T$ and $\mathcal{T}_S \neq \mathcal{T}_T$ and are thus transfer learning problems. However, the digit classes are contained in both \textit{MNIST} and the full \textit{NIST} datasets. Hence, the label space for \textit{MNIST} is a subset of the label space for \textit{NIST}, i.e. $\mathcal{Y}_{\text{MNIST}} \subset\mathcal{Y}_{\text{NIST}} $. This differentiates experiment sets 2 and 3, where $\mathcal{Y}_S \subset \mathcal{Y}_T$ and $\mathcal{Y}_S \cap \mathcal{Y}_T = \emptyset$ respectively.

For each of the experiment sets given above we investigate the behaviour for the \textit{target task control model}, the \textit{frozen weight transfer} and the \textit{fine-tuned transfer}. The \textit{target task control model} corresponds to a network trained on $\mathcal{T}_T$ with random weight initialisation. \textit{Frozen weight transfer} is achieved by allowing only the parameters of the final layer of the pre-trained model to be learnt; all other parameters are kept fixed. \textit{Fine-tuned transfer} is where all the parameters of the pre-trained model are allowed to be re-trained on $\mathcal{T}_T$.

For transfer to be performed, models need to be pre-trained on the source tasks. We pre-train networks on \textit{MNIST} and \textit{NIST Digits}, save their parameters, and use these as the initial parameters for the respective transfer models. Thirty CNNs are trained for each of these transfer types with various metrics logged after each training epoch. The networks are trained for up to 20 epochs, or where the early stopping criteria is met. The loss function $\mathcal{L}(\boldsymbol\theta)$ used is the \textit{Cross-Entropy Loss}, and the early stopping criteria is $\big|\mathcal{L}_{\text{test}}^{(k)} - \mathcal{L}_{\text{test}}^{(k-1)}\big| < 10^{-5}$, where $k$ is the epoch number. This means the network will stop training early when only a marginal improvement is made on the out-of-sample (test) loss.

The network architecture, optimiser and learning rates are kept constant throughout all experiments: the optimiser used is \textit{Adam}\cite{adam}, and a learning rate, $\alpha = 0.001$, is selected as informed by empirical tests and previous literature \cite{ciresandeepbig}.

The \textit{in-sample (training) loss}, \textit{out-of-sample loss}, and the \textit{out-of-sample F1-scores} are logged at every epoch to quantify convergence as well as model performance. We use F1-scores instead of accuracy scores due to class imbalance in NIST, where there are substantially more digit samples than character samples. For more detailed insight into the specific features that are transferred, we construct confusion matrices and per-class accuracy plots for each experiment set.

\section{Results and discussion} \label{results}
\subsection{MNIST to NIST Digits}

\begin{figure}[b]
    \centering
    \includegraphics[width=\columnwidth]{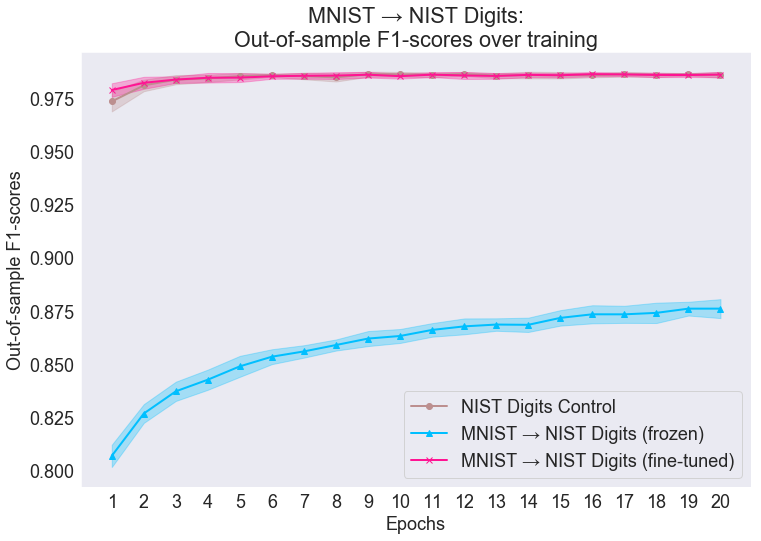}
    \caption{Per epoch out-of-sample F1-scores for  \textit{MNIST} to \textit{NIST Digits} transfer.}
    \label{fig:mnist-nist-dig-f1s}
\end{figure}

For this set of experiments, we observe that the \textit{fine-tuned transfer} and the \textit{control} networks perform very well on the \textit{NIST Digits} classification (Fig. \ref{fig:mnist-nist-dig-f1s}). We note that the \textit{fine-tuned} transfer has slightly better out-of-sample performance for the first two epochs of training as compared with the \textit{control} networks, but both models perform the same thereafter. After this 2 epoch point, both these models converge and plateau at a mean F1-score of approximately $0.98$.
This result suggests that transferring parameters provides a slightly better starting point for the network when compared to using randomly initialised parameters for this problem, and thus converges in fewer epochs than the \textit{control} model. However, for more than 2 epochs of training, no improvement in performance is gained by using \textit{fine-tuned transfer}.

\textit{Frozen-weight transfer} performs worse than both the \textit{control} and \textit{fine-tuned} networks. This demonstrates that the features transferred from \textit{MNIST} are able to classify a large portion of the \textit{NIST Digits}, but these features cannot capture the variation in the target task as well as the \textit{fine-tuned} network can. It is possible that upscaling the MNIST images hinders the \textit{frozen} transfer performance, and it remains to be seen whether an improvement can be made by downscaling the images instead.  The \textit{frozen-weight} performance keeps improving over time and does not plateau within the 20-epoch training period. Despite the truncated training period, the gradient of the \textit{frozen} network performance indicates that the network will not reach comparable performance to the other two transfer types in a reasonable amount of training time. Per-class accuracy analysis of the \textit{frozen-weight transfer} demonstrate that the network performs consistently worse on certain classes throughout its training.

\subsection{MNIST to NIST Digits + Characters}
\begin{figure}[b]
    \centering
    \includegraphics[width=\columnwidth]{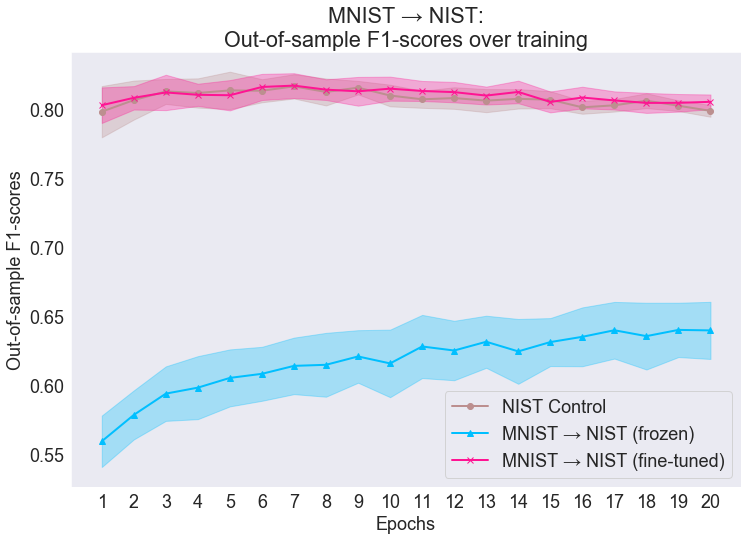}
    \caption{Per epoch out-of-sample F1-scores for  \textit{MNIST} to \textit{NIST Digits + Characters} transfer.}
    \label{fig:mnist-nist-f1s}
\end{figure}

As in the previous experiment, we observe that the shapes of both the \textit{control} networks and \textit{fine-tuned transfer} graphs are quite similar. However for this problem the networks do not plateau: instead they reach their respective peaks at around 6 epochs, and their performance deteriorates gradually thereafter (see Fig. \ref{fig:mnist-nist-f1s}). Initially, the \textit{fine-tuned} network performs better than the \textit{control} model, and we note with less variance in its performance. This implies that transferring parameters provides a moderately better starting point on the full \textit{NIST} set more regularly than random parameter initialisation does.  Both networks achieve peak mean F1-scores just above $0.82$, indicating no significant performance gains are achieved by undertaking parameter transfer.

\textit{Frozen-weight transfer} again performs more poorly than the other two networks. This is expected as the domains and tasks are now different, and the features learnt on the digits may not be useful in discriminating between the combined characters and digits. The \textit{frozen-weight} network performance improves with more training, with a similar shape to that seen in the previous experiment set. Thus we can draw the same conclusion: that the \textit{frozen} network will not reach similar performance to the \textit{fine-tuned} and \textit{control} networks in a reasonable amount of training time.

\begin{figure}[t]
    \centering
    \begin{subfigure}[b]{\columnwidth}
        \includegraphics[width=1\linewidth]{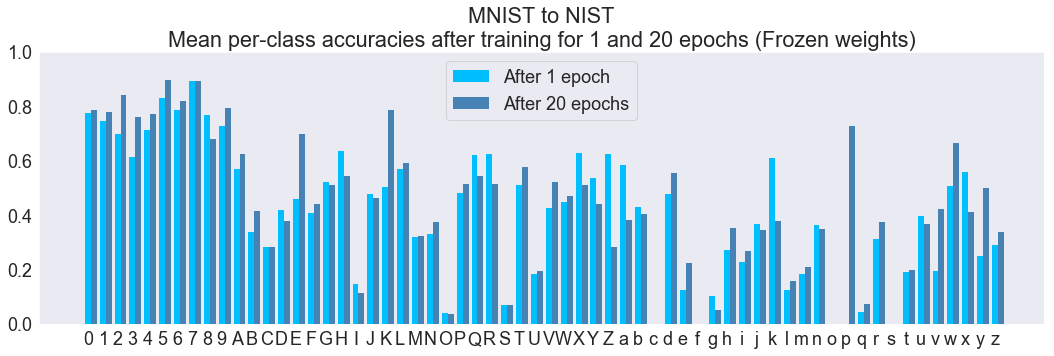}
        \caption{\textit{Frozen-weight} transfer}
        \label{fig:mnist-nist-per-class-freeze}
    \end{subfigure}
    
    \begin{subfigure}[b]{\columnwidth}
        \includegraphics[width=1\linewidth]{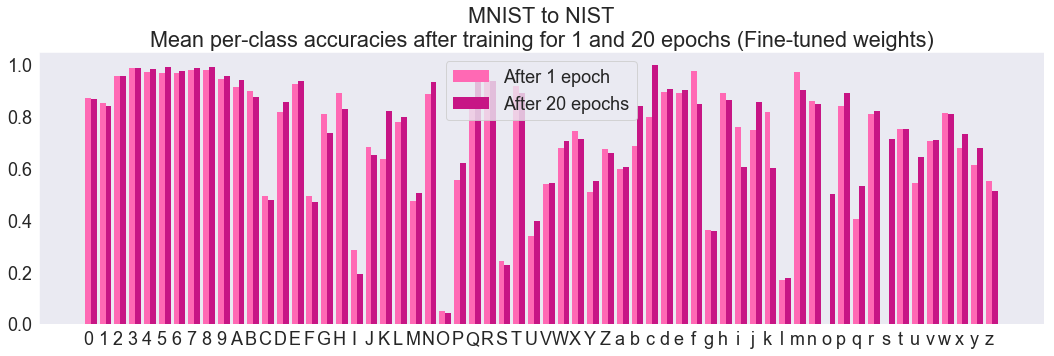}
        \caption{\textit{Fine-tuned} transfer}
        \label{fig:mnist-nist-per-class-ft}
    \end{subfigure}
    
    \caption{Mean per-class accuracy scores after 1 and 20 epochs for \textit{MNIST} to \textit{NIST Digits + Characters}.}
    \label{fig:mnist-nist-per-class}
\end{figure}

After plotting per-class accuracy scores across training, we can visualise how well specific features are transferring. From  this analysis, we observe that the \textit{frozen} network performs much better on the digits than on the characters. After a single epoch of training, the \textit{frozen-weight} networks incorrectly classify all samples of the lowercase letters ``c'', ``f'', ``g'', ``m'', ``o'', ``p'', ``s'', ``u'' and ``v''. This is probably due to these characters being spatially similar to their respective uppercase letters. After more epochs of training the \textit{frozen} networks, not much improvement is made for these classes (Fig \ref{fig:mnist-nist-per-class-freeze}), indicating that the features transferred from \textit{MNIST} are not useful in classifying the lowercase characters on their own. The per-class accuracy scores for the \textit{fine-tuned} networks show a marked improvement for the corresponding classes' performance after re-training the parameters (Fig \ref{fig:mnist-nist-per-class-ft}).

\subsection{NIST Digits to NIST Characters}
\begin{figure}[t]
    \centering
    \includegraphics[width=\columnwidth]{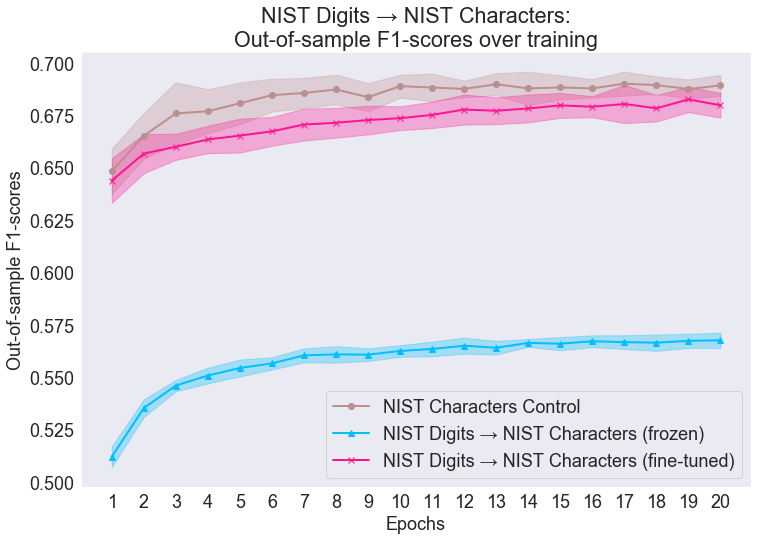}
    \caption{Per epoch out-of-sample F1-scores for  \textit{NIST Digits} to \textit{NIST Characters} transfer.}
    \label{fig:nist-dig-char-f1s}
\end{figure}

In this experiment, the \textit{fine-tuned transfer} performs slightly worse than the \textit{control} networks do (see Fig. \ref{fig:nist-dig-char-f1s}). This demonstrates that the parameters transferred provide a worse starting point than random initialisation does for classifying \textit{NIST} characters. This is quite different to the behaviour seen in the other experiment sets, where the \textit{control} and \textit{fine-tune} transfer perform comparably. This is possibly due to the differences in label spaces between this task and the other two tasks: i.e. here we have $\mathcal{Y}_S \not\subset \mathcal{Y}_T$ whereas the other sets have $\mathcal{Y}_S \subseteq \mathcal{Y}_T$.

The \textit{frozen} networks once again perform much worse than the \textit{fine-tune} and \textit{control} networks do. We observe a similar peak F1-error rate of roughly 42\% as Cireşan, \textit{et al}. \cite{ciresanchinese} for the identical transfer problem. The shape of the graph leads to the same deduction about performance after the 20 epoch cut-off period as for the previous \textit{frozen} networks.  Per-class accuracy analysis shows that the network performs badly on certain classes (e.g. ``o'', ``c''), then later improve on these classes but worsen on other classes. This result demonstrates that transferring the features learnt for \textit{NIST Digits} is not particularly beneficial for discriminating between the 52 classes of characters.

Overall, all three transfer types perform more poorly on \textit{NIST Characters} than for the full \textit{NIST} set. This is due to the networks performing better on the digits and thus improving the overall F1-score for the full \textit{NIST} dataset. This is seen in per-class analysis and noted in related research \cite{ciresanmulticolumn}. 

\subsection{Summary of results}
\begin{figure}[htb]
    \centering
    \includegraphics[width=\columnwidth]{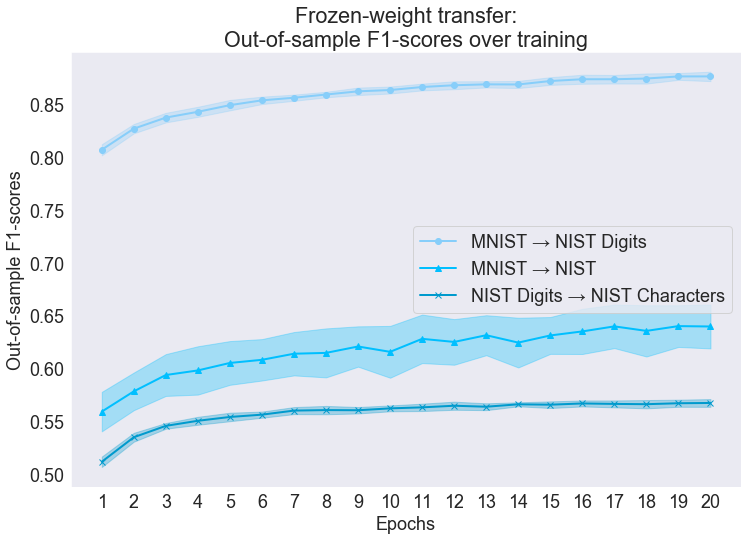}
    \caption{\textit{Frozen weight} transfer across training for all experiment sets.}
    \label{fig:frozen}
\end{figure}

Overall we observe that transferring features (\textit{frozen-weight transfer}) performs significantly worse than both randomly initialised networks and parameter transfer (\textit{fine-tuned transfer}) across all experiment sets. This indicates that the features learnt on $\mathcal{T}_S$ are not complex enough to discriminate between classes in $\mathcal{T}_T$, even for the case where $\mathcal{T}_S = \mathcal{T}_T$. Despite this, \textit{frozen} transfer for \textit{MNIST} to \textit{NIST Digits} achieves substantially better F1-scores than for the other \textit{frozen} networks (Fig. \ref{fig:frozen}, due to the overlap in source and target tasks. Random initialisation predicts an upper bound for \textit{fine-tuned} transfer performance in all three sets. The latter provides improved convergence in certain tasks but with diminishing performance gains after more training. Per-class accuracy analysis over training demonstrates that feature transfer is limited by the features it learns on $\mathcal{T}_S$, whereas \textit{fine-tuning} the networks allow these feature maps to be tweaked to perform better on $\mathcal{T}_T$.

In all our experiment sets, we find that a transfer learning approach provides no significant benefits for learning the target tasks, both for network performance and its convergence. This is contrary to the results demonstrated for transfer learning using ImageNet where improved accuracy and model convergence were achieved \cite{betterImagenet, imagenet_good, howtransferable}. This suggests that by itself, transfer learning (and specifically deep transfer learning) does not necessarily imply an improved or more robust model, at least within the domains we perform our experiments.
\section{Conclusion} \label{conclusion}
In this paper, we set up experiments to investigate properties of different types of knowledge transfer, specifically for character recognition tasks. We compare randomly initialised networks to both \textit{frozen-weight} networks (transferring features) and \textit{fine-tuned} networks (transferring parameters). We demonstrate that the randomly initialised networks perform better or similarly to the \textit{fine-tuned} networks on target tasks, and thus no improvements in performance are achieved. \textit{Frozen-weight transfer} performs considerably worse than the other types of transfer for all sets of experiments.

Our results demonstrate that deep transfer learning does not provide any substantial benefit when compared to traditional deep learning for the character recognition problems we pose. Amid many recent successes in transfer learning, especially with ImageNet, our work provides cases where knowledge transfer is not particularly useful. Thus, contrary to the notion that knowledge transfer necessarily implies better network performance or convergence, we provide counterexamples where this is not the case.

\bibliography{refs}

\begin{thebibliography}{10}
\providecommand{\url}[1]{#1}
\csname url@samestyle\endcsname
\providecommand{\newblock}{\relax}
\providecommand{\bibinfo}[2]{#2}
\providecommand{\BIBentrySTDinterwordspacing}{\spaceskip=0pt\relax}
\providecommand{\BIBentryALTinterwordstretchfactor}{4}
\providecommand{\BIBentryALTinterwordspacing}{\spaceskip=\fontdimen2\font plus
\BIBentryALTinterwordstretchfactor\fontdimen3\font minus
  \fontdimen4\font\relax}
\providecommand{\BIBforeignlanguage}[2]{{%
\expandafter\ifx\csname l@#1\endcsname\relax
\typeout{** WARNING: IEEEtran.bst: No hyphenation pattern has been}%
\typeout{** loaded for the language `#1'. Using the pattern for}%
\typeout{** the default language instead.}%
\else
\language=\csname l@#1\endcsname
\fi
#2}}
\providecommand{\BIBdecl}{\relax}
\BIBdecl

\bibitem{alexnet}
A.~Krizhevsky, I.~Sutskever, and G.~E. Hinton, ``Imagenet classification with
  deep convolutional neural networks,'' in \emph{Advances in neural information
  processing systems}, 2012, pp. 1097--1105.

\bibitem{googlenet}
C.~Szegedy, W.~Liu, Y.~Jia, P.~Sermanet, S.~Reed, D.~Anguelov, D.~Erhan,
  V.~Vanhoucke, and A.~Rabinovich, ``Going deeper with convolutions,'' in
  \emph{Proceedings of the IEEE conference on computer vision and pattern
  recognition}, 2015, pp. 1--9.

\bibitem{ciresandeepbig}
D.~C. Cire{\c{s}}an, U.~Meier, L.~M. Gambardella, and J.~Schmidhuber, ``Deep,
  big, simple neural nets for handwritten digit recognition,'' \emph{Neural
  computation}, vol.~22, no.~12, pp. 3207--3220, 2010.

\bibitem{betterImagenet}
S.~Kornblith, J.~Shlens, and Q.~V. Le, ``Do better {ImageNet} models transfer
  better?'' in \emph{Proceedings of the IEEE Conference on Computer Vision and
  Pattern Recognition}, 2019, pp. 2661--2671.

\bibitem{conv}
C.~Nebauer, ``Evaluation of convolutional neural networks for visual
  recognition,'' \emph{IEEE Transactions on Neural Networks}, vol.~9, no.~4,
  pp. 685--696, 1998.

\bibitem{imagenet_good}
M.~Huh, P.~Agrawal, and A.~A. Efros, ``What makes {ImageNet} good for transfer
  learning?'' in \emph{NIPS Large Scale Computer Vision Systems Workshop
  (Oral)}, 2016.

\bibitem{mnist}
Y.~LeCun, C.~Cortes, and C.~J. Burges, ``The {MNIST} database of handwritten
  digits, 1998,'' \emph{URL http://yann.lecun.com/exdb/mnist}, 1998.

\bibitem{nist}
P.~J. Grother, ``{NIST} special database 19,'' \emph{Handprinted forms and
  characters database, National Institute of Standards and Technology}, 1995.

\bibitem{panyang}
S.~J. Pan and Q.~Yang, ``A survey on transfer learning,'' \emph{IEEE
  Transactions on Knowledge and Data Engineering}, vol.~22, no.~10, pp.
  1345--1359, 2009.

\bibitem{thrunclustering}
S.~Thrun and J.~O’Sullivan, ``Clustering learning tasks and the selective
  cross-task transfer of knowledge,'' in \emph{Learning to learn}.\hskip 1em
  plus 0.5em minus 0.4em\relax Springer, 1998, pp. 235--257.

\bibitem{haykin-nn}
S.~S. Haykin \emph{et~al.}, \emph{Neural networks and learning machines},
  3rd~ed.\hskip 1em plus 0.5em minus 0.4em\relax New York: Prentice Hall, 2009.

\bibitem{bengiodeep}
Y.~Bengio, ``Deep learning of representations for unsupervised and transfer
  learning,'' in \emph{Proceedings of ICML workshop on unsupervised and
  transfer learning}, 2012, pp. 17--36.

\bibitem{howtransferable}
J.~Yosinski, J.~Clune, Y.~Bengio, and H.~Lipson, ``How transferable are
  features in deep neural networks?'' in \emph{Advances in Neural Information
  Processing Systems}, 2014, pp. 3320--3328.

\bibitem{imagenet}
J.~Deng, W.~Dong, R.~Socher, L.-J. Li, K.~Li, and L.~Fei-Fei, ``{ImageNet}: A
  large-scale hierarchical image database,'' in \emph{IEEE conference on
  Computer Vision and Pattern Recognition}.\hskip 1em plus 0.5em minus
  0.4em\relax IEEE, 2009, pp. 248--255.

\bibitem{ciresanchinese}
D.~C. Cire{\c{s}}an, U.~Meier, and J.~Schmidhuber, ``Transfer learning for
  {L}atin and {C}hinese characters with deep neural networks,'' in \emph{The
  2012 International Joint Conference on Neural Networks (IJCNN)}.\hskip 1em
  plus 0.5em minus 0.4em\relax IEEE, 2012, pp. 1--6.

\bibitem{resnet}
K.~He, X.~Zhang, S.~Ren, and J.~Sun, ``Deep residual learning for image
  recognition,'' in \emph{Proceedings of the IEEE Conference on Computer Vision
  and Pattern Recognition}, 2016, pp. 770--778.

\bibitem{adam}
D.~P. Kingma and J.~Ba, ``Adam: A method for stochastic optimization,''
  \emph{arXiv preprint arXiv:1412.6980}, 2014.

\bibitem{ciresanmulticolumn}
D.~Cire{\c{s}}an, U.~Meier, and J.~Schmidhuber, ``Multi-column deep neural
  networks for image classification,'' in \emph{2012 IEEE Conference on
  Computer Vision and Pattern Recognition}, 2012.

\end{thebibliography}

\end{document}